\documentclass[english]{lni}
\usepackage[utf8]{inputenc}
\usepackage{subcaption}

\newcommand{\px}{$\:$px }
\newcommand{\pxns}{$\:$px}

\let\subsectionautorefname\sectionautorefname
\let\subsubsectionautorefname\sectionautorefname

\begin{document}
%%% Mehrere Autoren werden durch \and voneinander getrennt.
%%% Die Fußnote enthält die Adresse sowie eine E-Mail-Adresse.
%%% Das optionale Argument (sofern angegeben) wird für die Kopfzeile verwendet.
\title[Automatic Plant Cover Estimation with CNNs]{Automatic Plant Cover Estimation with Convolutional Neural Networks}
%%%\subtitle{Untertitel / Subtitle} % if needed
\author[Körschens et al.]
{
    Matthias Körschens \footnote{Computer Vision Group, Friedrich Schiller University Jena, 07737 Jena, Germany; \email{matthias.koerschens@uni-jena.de}}
    \footnote{Plant Biodiversity Group, Friedrich Schiller University Jena, 07737 Jena, Germany}
    \and%
    Paul Bodesheim \footnote{Computer Vision Group, Friedrich Schiller University Jena, 07737 Jena, Germany; \email{paul.bodesheim@uni-jena.de}} \and%
    Christine Römermann \footnote{Plant Biodiversity Group, Friedrich Schiller University Jena, 07737 Jena, Germany; \email{christine.roemermann@uni-jena.de}} \and%
    Solveig Franziska Bucher \footnote{Plant Biodiversity Group, Friedrich Schiller University Jena, 07737 Jena, Germany; \email{solveig.franziska.bucher@uni-jena.de}} \and%
    Mirco Migliavacca \footnote{Department Biogeochemical Integration, Max Planck Institute for Biogeochemistry, 07701 Jena, Germany; \email{mmiglia@bgc-jena.mpg.de}} \and%
    Josephine Ulrich \footnote{Plant Biodiversity Group, Friedrich Schiller University Jena, 07737 Jena, Germany; \email{josephine.ulrich@uni-jena.de}} \and%
    Joachim Denzler \footnote{Computer Vision Group, Friedrich Schiller University Jena, 07737 Jena, Germany; \email{joachim.denzler@uni-jena.de}} 
}

\startpage{1} % Beginn der Seitenzählung für diesen Beitrag / Start page
\editor{Gesellschaft für Informatik e.V. (GI)} % Names of Editors
\booktitle{INFORMATIK 2021} % Name of book title
\yearofpublication{2021}
%%%\lnidoi{18.18420/provided-by-editor-02} % if known
\maketitle

\begin{abstract}%70-150 words
    Monitoring the responses of plants to environmental changes is essential for plant biodiversity research. This, however, is currently still being done manually by botanists in the field. This work is very laborious, and the data obtained is, though following a standardized method to estimate plant coverage, usually subjective and has a coarse temporal resolution. To remedy these caveats, we investigate approaches using convolutional neural networks (CNNs) to automatically extract the relevant data from images, focusing on plant community composition and species coverages of 9 herbaceous plant species. To this end, we investigate several standard CNN architectures and different pretraining methods. We find that we outperform our previous approach at higher image resolutions using a custom CNN with a mean absolute error of 5.16\%. In addition to these investigations, we also conduct an error analysis based on the temporal aspect of the plant cover images. This analysis gives insight into where problems for automatic approaches lie, like occlusion and likely misclassifications caused by temporal changes.
    % Maybe explain the last part further? Word limit for abstract is already exceeded...
\end{abstract}
%%% Beginn des Artikeltexts
\section{Introduction}

Environmental changes can have a large impact not only on human life, but also on animals and plants, and can have drastic effects on biodiversity. This gives reason to steadily monitor nature for such changes and their impacts. Plants, for example, are strong indicators of environmental changes like climate change. This can be seen in phenological responses \cite{rosenzweig2007assessment,menzel2006european,bucher2018traits,miller2008global,cleland2012phenological,fitter2002rapid}, and also in changes in plant community compositions \cite{rosenzweig2007assessment,liu2018shifting,lloret2009plant}. However, even more environmental aspects can be monitored through plants, like land use \cite{gerstner2014landuse,aggemyr2012landscape} and insect abundance \cite{Souza2016BottomupAT}.
Therefore, plant community compositions are the focus of a large number of experiments \cite{liu2018shifting,gerstner2014landuse,Souza2016BottomupAT,bruelheide2018global}, where the composition is usually recorded at regular time intervals. This is usually done by estimating the plant cover of each species in the predefined vegetation plots. The cover of a plant species describes the percentage of soil area covered by the respective species. In many experiments, this value is estimated at intervals of at least about a week, usually much longer, due to it being a very laborious task, and the estimation itself is usually done by a single biologist. This introduces two problems: a low temporal resolution of recordings of the plant community composition and noisy data caused by human error and subjectivity of the estimations.
These problems can be mitigated by automatic image analysis methods. However, while in current times there are many possibilities to collect images and videos of vegetation plots in an automated manner (e.g., \cite{brown2016using}), performing more complicated analyses automatically like the cover estimation mentioned above is still not possible as we are still missing the methods to do so. 

Convolutional neural networks (CNNs) have proven to be a very effective means for image analysis in general and hence could also be utilized as a basis for such methods. However, automatic plant cover prediction is a rather complicated task, as it usually includes large images containing many plant species with only subtle differences, which strongly occlude each other. In addition to this, the annotations for plant cover are relative values (percentages), which is a mostly unexplored kind of annotation in the area of computer vision.
In our previous work \cite{koerschens2020towards} we laid the groundwork to solve the task of plant cover prediction. 

In this work, we will investigate that approach, i.e., the cover prediction via pixel-wise classification using the previously proposed network head, in more detail regarding multiple aspects. Firstly, we aim to investigate how different standard CNN architectures with differing characteristics perform on our task in comparison to the network from \cite{koerschens2020towards}. The architectures we compare are a ResNet50 \cite{he2015resnet}, InceptionV3 \cite{szegedy2015inceptionv3} and DenseNet121 \cite{iandola2014densenet}. Secondly, we investigate whether the addition of information from earlier network layers of the abovementioned networks and an additional increase of the output resolution by using a Feature Pyramid Network \cite{lin2017feature} can improve the performance in comparison to the networks' standalone performance. Thirdly, as changing the image resolution during training and evaluation can strongly affect the network performance, for example, due to the limitations of the networks' receptive field size, we evaluate the performance of the abovementioned CNN architectures on different image resolutions. Fourthly, as pretraining in a similar domain to the target task usually results in better network performances than out-of-domain pretraining \cite{cui2018large}, we aim to find out how large the benefits of this method are for plant cover prediction. To this end, we compare the previously mentioned configurations in two different pretraining settings. For our domain-similar (``in-domain'') pretraining setting, we constructed a dedicated plant image dataset using images from the GBIF website\footnote{Global Biodiversity Information Facility, \url{https://www.gbif.org/}}, here referred to as the GBIF dataset. In our out-of-domain setting, we utilize freely available off-the-shelf Imagenet \cite{russakovsky2015imagenet} pretraining weights. Lastly, to explain the errors made by the networks during the plant cover prediction, we conduct a temporal analysis of the prediction error of a selected network based on the week of recording of the respective images. 

In the following, we will elaborate on related work in the area of automatic plant analysis (\autoref{sec:related_work}), followed by an introduction of the datasets and methods we use in \autoref{sec:datasets} and \autoref{sec:method}, respectively. After that, we present the results of our experiments (\autoref{sec:experiments}) and end the paper with a conclusion and an outlook on possible future work.

%In the remainder of this paper we will elaborate on related work in the area of automatic plant identification in general and plant cover prediction in particular. This is followed by an explanation of the details and intricacies of the dataset and consecutively the method we use. In the experimental part in \autoref{sec:experiments} we will initially show a comparison of different network architectures used to solve the task of plant cover determination, followed by an ablation study of the effect of pretraining on a related dataset in comparison to ImageNet pretraining. We will moreover also analyse how the prediction error behaves depending on the week the images were taken.

% More citations, outline, contributions

\section{Related Work}
\label{sec:related_work}

%In this section we will summarize already existing work in the area of plant analysis and plant cover prediction in particular.

\subsection{Plant Identification}

In the last years, there have been several CNN-based approaches developed for the identification of plant species from images \cite{yalcin2016plant,barre2017leafnet,lee2015deep,ghazi2017plant}. Barré \etal \cite{barre2017leafnet} developed a rather simple CNN architecture called LeafNet, whose aim is to determine the plant species from leaf images. Yalcin \etal \cite{yalcin2016plant} utilize a pretrained CNN with 11 layers to classify images of agricultural plants. However, simple identification of plant species via images is not the only area where Deep Learning approaches like CNNs are applied. Other tasks include the detection of fruits via segmentations \cite{ganesh2019deep}, counting of fruits of agricultural plants via regression networks \cite{lu2017tasselnet,xiong2019tasselnetv2} and plant disease prediction \cite{chen2020using}. A more comprehensive list of approaches can be found, for example, in the survey by Kamilaris \etal \cite{kamilaris2018deep}.

% Maybe mention why not applicable for PCP?

\subsection{Plant Cover Prediction}

While the number of approaches for problems concerning plant identification is rather abundant, only two approaches try to solve the problem of cover prediction via images. The first one is by Kattenborn \etal \cite{kattenborn2020convolutional}, who tried to determine the cover percentages of woody species from UAV-based remote sensing data. For their analysis, they utilized a simple custom CNN with eight layers to calculate the cover percentages for several herb, shrub and woody species. In contrast to the data used in this paper, there was not only no occlusion to be taken into account, but the annotations were delineations in the images. In addition to that, the woody plant species investigated mostly had heterogeneous appearances, making automatic discrimination between them relatively easy compared to the much more similar herbaceous species found in the InsectArmageddon dataset \cite{ulrich2020invertebrate,koerschens2020towards} we use.
The second approach in the area of plant cover prediction is our recent work presented in \cite{koerschens2020towards}, which we will build on in this paper. In \cite{koerschens2020towards} we utilized a novel custom network architecture with 12 layers and a block similar to those of the Inception network \cite{szegedy2015inceptionv3} for feature aggregation. The respective experiments were also done on the InsectArmageddon dataset from Ulrich \etal \cite{ulrich2020invertebrate}. In contrast to our previous work, we will not utilize a custom network here but evaluate the performance of several standard CNN architectures in conjunction with Feature Pyramid Networks (FPNs) \cite{lin2017feature}. To this end, we utilize the network head proposed in \cite{koerschens2020towards} and hence adopt the pixel-wise classification approach. This way, we aim to determine how well these architectures are suited for this task and if the shortcut connections and higher output resolution added by the FPN can be beneficial for plant cover prediction. In addition to this, we also investigate different image resolutions as well as different types of pretraining, as they can strongly affect the performance of CNNs.

\section{Datasets}
\label{sec:datasets}

\newcommand{\pimgsize}{0.16\textwidth}

\begin{figure}[t]
    \centering
    \begin{minipage}{\linewidth}
        \centering
        \includegraphics[width=\pimgsize]{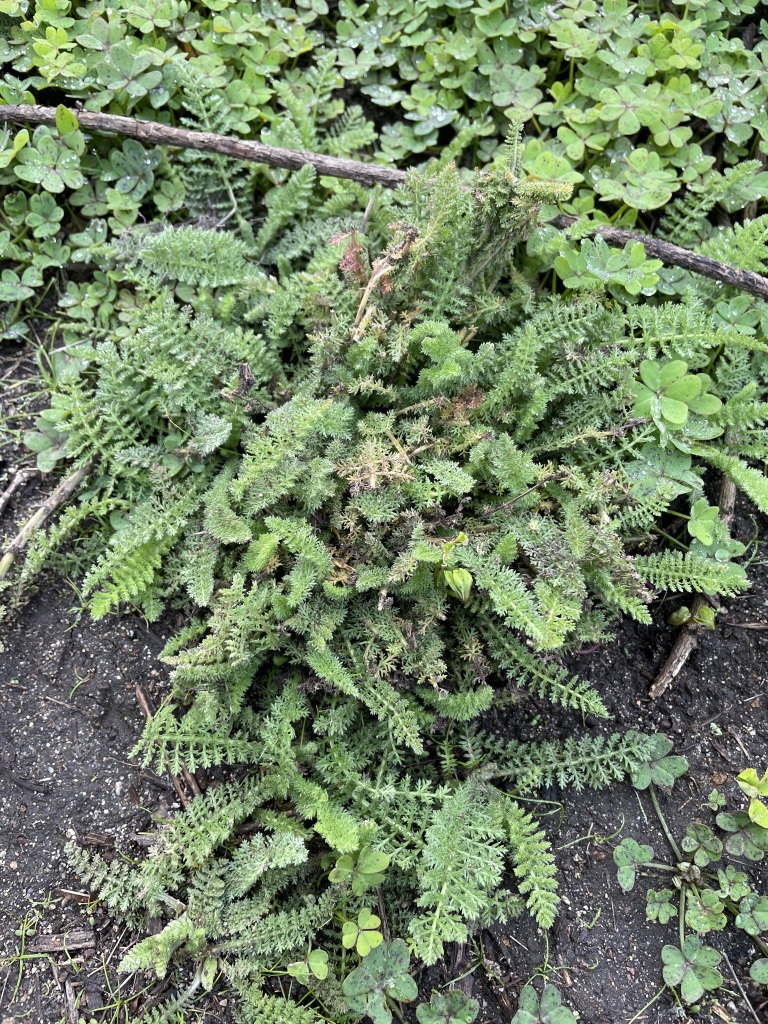}
        \includegraphics[width=\pimgsize]{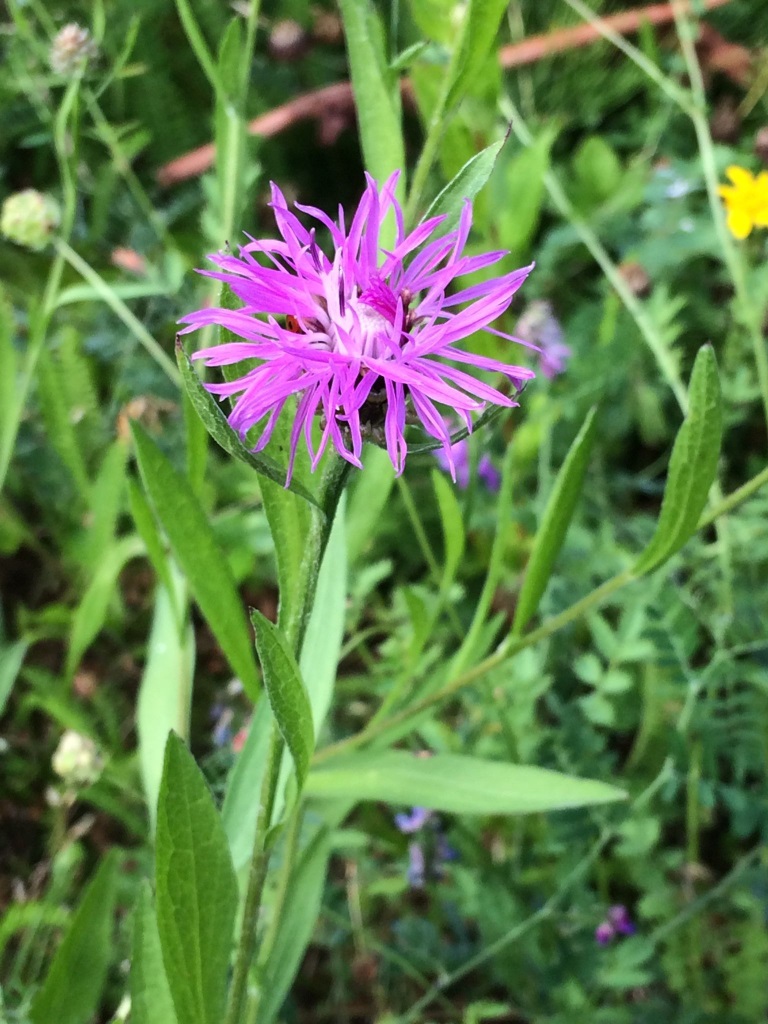}
        \includegraphics[width=\pimgsize]{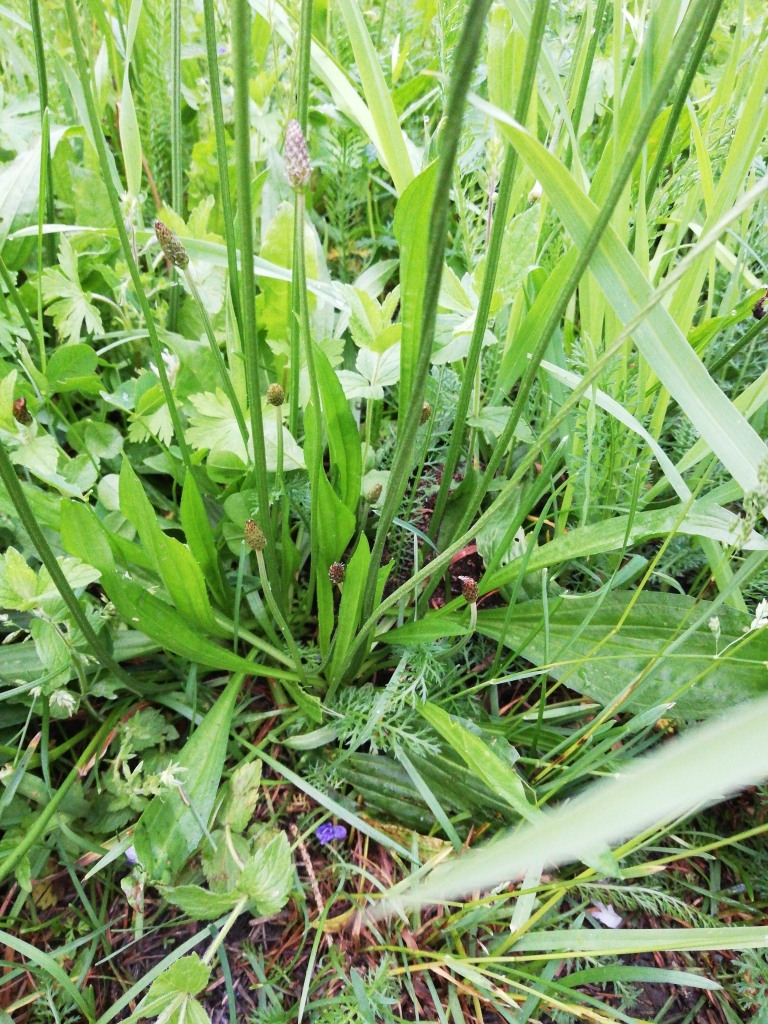}
        \includegraphics[width=\pimgsize]{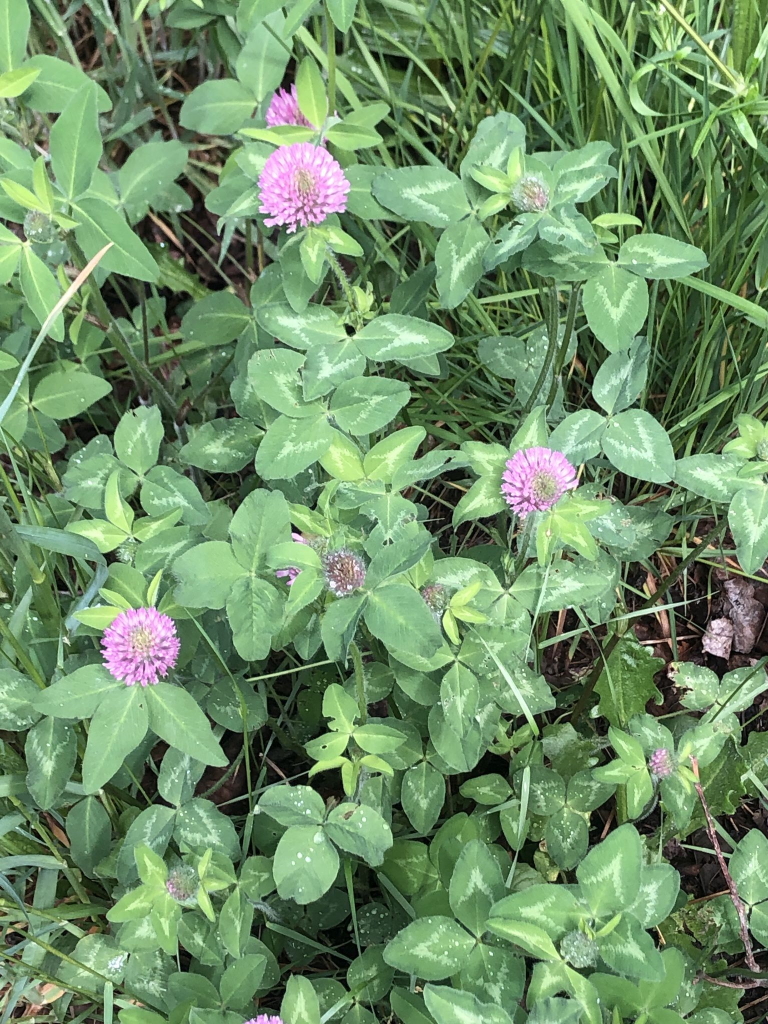}
        \includegraphics[width=\pimgsize]{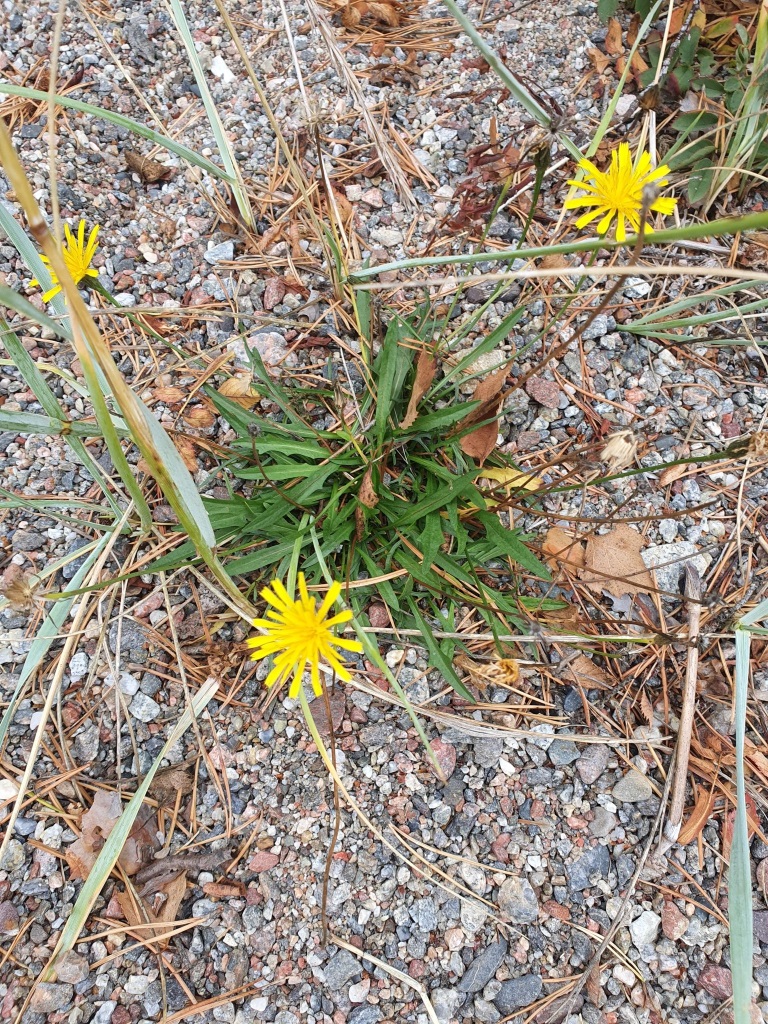}
        \includegraphics[width=\pimgsize]{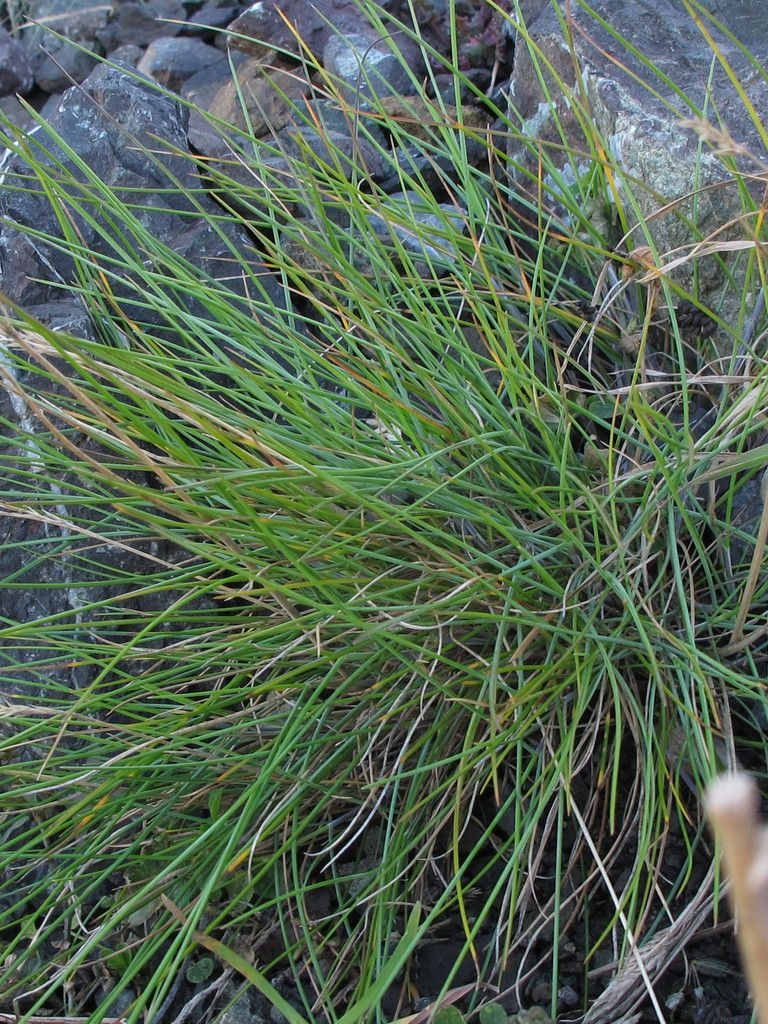}
    \end{minipage}
    
    % \begin{minipage}{\linewidth}
    %     \centering
    % \end{minipage}
    \caption{Example images from our GBIF dataset. The plant species shown are from left to right: \textit{Achillea millefolium} (Yarrow), \textit{Centaurea jacea} (brown knapweed), \textit{Plantago lanceolata} (ribwort plantain), \textit{Trifolium pratense} (red clover), \textit{Scorzoneroides autumnalis} (autumn hawkbit) and \textit{Grasses}, which are not differentiated into different species.}
    \label{fig:examples_gbif}
\end{figure}

In our experiments, we utilize two datasets. The first one is used for pretraining and will be referred to as GBIF dataset, as it contains plant images taken from the Global Biodiversity Information Facility (GBIF)\footnote{\url{https://www.gbif.org/}}. The second dataset is the so-called InsectArmageddon dataset introduced in \cite{ulrich2020invertebrate} and \cite{koerschens2020towards}. These datasets will be explained in \autoref{subsec:gbif} and \autoref{subsec:ia_ds}, respectively.

\subsection{The GBIF Dataset}
\label{subsec:gbif}

The GBIF dataset is a classification dataset we designed specifically for pretraining models in preparation for using them for cover prediction on the InsectArmageddon dataset. Hence, it contains mostly the same classes as the latter, namely the plant species \textit{Achillea millefolium} (Ach\_mil), \textit{Centaurea jacea} (Cen\_jac), \textit{Lotus corniculatus} (Lot\_cor), \textit{Medicago lupulina} (Med\_lup), \textit{Plantago lanceolata} (Pla\_lan), \textit{Scorzoneroides autumnalis} (Sco\_aut) and \textit{Trifolium pratense} (Tri\_pra), as well as \textit{Festuca Tourn. ex L.}, which represents the collective class of grasses in the InsectArmageddon dataset, in which the individuals are not further identified to the species-level.

This dataset contains $7{,}200$ images with 750 training and 150 validation images per class and is therefore also balanced. The images were selected randomly from the complete sets of human observations of the respective plant species \cite{gbif2020occurrences}. A selection of example images from the dataset is shown in \autoref{fig:examples_gbif}.

\subsection{The InsectArmageddon Dataset}
\label{subsec:ia_ds}

\begin{figure}[t]
    \centering
    \includegraphics[width=\textwidth]{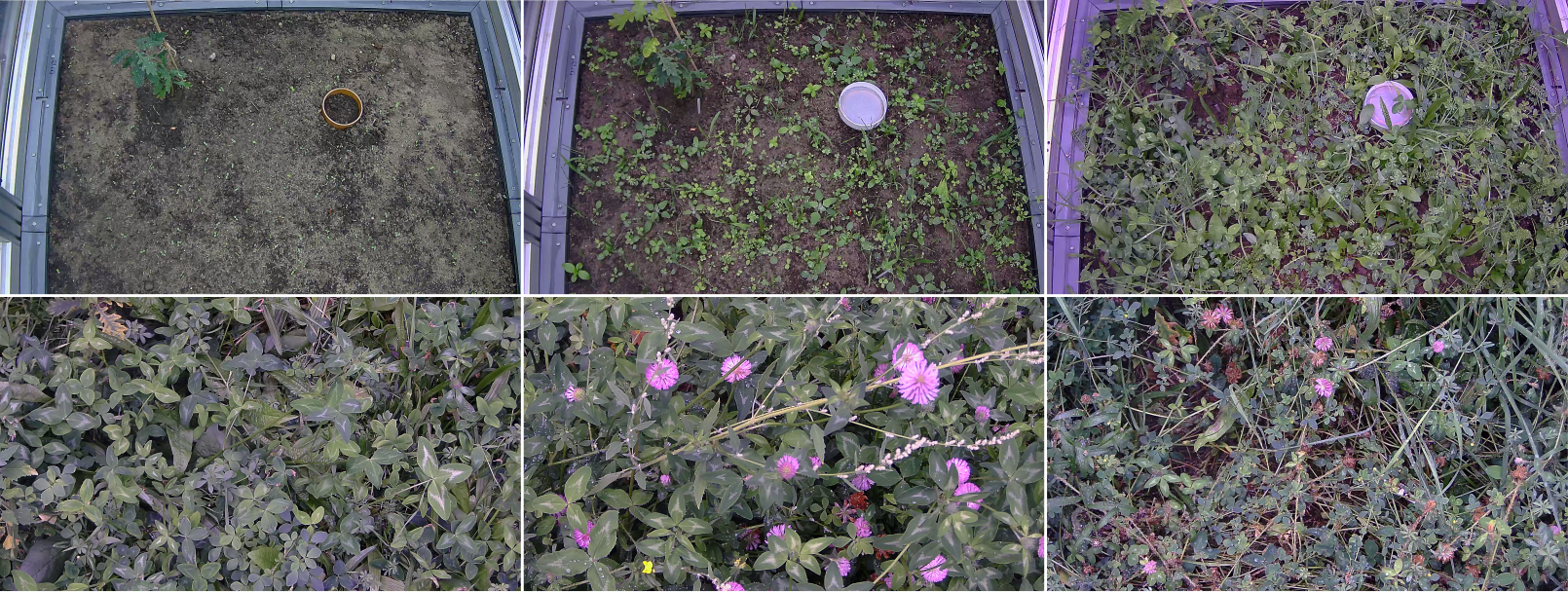}
    \caption{Example images from the InsectArmageddon dataset. We can see temporal changes in plant coverages in the experimental units of the Ecotron, where different plant species have been grown from seed.}
    \label{fig:examples_ia}
\end{figure}

\let\subsectionautorefname\sectionautorefname
\let\subsubsectionautorefname\sectionautorefname

The InsectArmageddon dataset was generated during the eponymous project\footnote{\url{https://www.idiv.de/en/research/platforms_and_networks/idiv_ecotron/experiments/insect_armageddon.html}} in 2018, in which the effect of invertebrate density on plant composition and growth were investigated \cite{ulrich2020invertebrate}. In this experiment, 24 so-called EcoUnits, i.e., enclosed boxes with a base area of about $1.5\:m\times1.5\:m$ containing small, closed ecosystems with different experimental treatments, were equipped with two cameras each, collecting images from a height of about 2$\:$m over a duration of 18 weeks. During this time, plants of 9 different species grew in the EcoUnits, resulting in large variations of the plants in the images as seen in \autoref{fig:examples_ia}. The plant species in this dataset are the same as in \autoref{subsec:gbif} with the addition of dead biomass, referred to as \textit{Dead litter}, which was introduced due to lack of visual distinguishability in the images. For more information on the InsectArmageddon experiment itself or the image collection process, we would like to refer to \cite{ulrich2020invertebrate,eisenhauer2018climate,turke2017multitrophische} and \cite{koerschens2020towards}, respectively. 

\begin{figure}[t]
    \centering
    \includegraphics[width=0.35\textwidth]{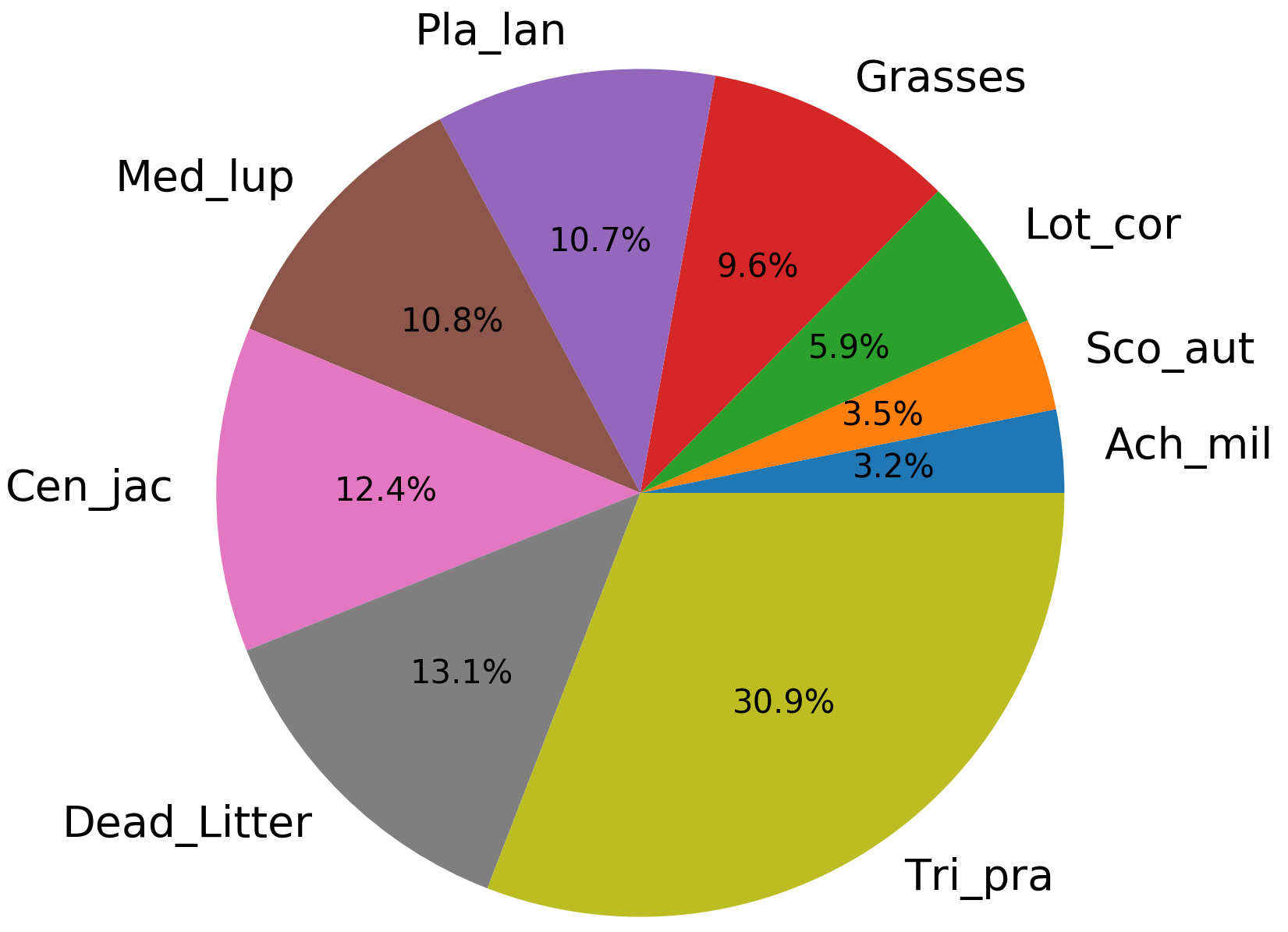}
    \caption{The distribution of plant species in the InsectArmageddon dataset, displayed as fractions of the total sum of cover percentages over the complete dataset. \textit{Trifolium pratense} is the most abundant plant in the dataset, while \textit{Achillea millefolium} represents the least abundant species.}
    \label{fig:distribution_ia}
\end{figure}

The images themselves contain the plant species in largely different distributions, partially caused by the different treatments. Therefore, the contents of the images can differ strongly. In addition to this, the dataset is heavily imbalanced. On the one hand, multiple plant species are either very abundant or have large leaf sizes, each leading to large cover percentages. On the other hand, there are also plant species with tiny leaves, leading to excessively small cover percentages. This distribution is visualized in \autoref{fig:distribution_ia}, which shows that the most prevalent plant is \textit{Trifolium pratense}, representing about a third of the complete dataset. In contrast, \textit{Achillea millefolium} can be found on the other end of the distribution with only about 3\% cover.

The dataset contains a large number of difficulties for automatic image analysis methods. Due to technical reasons, the zoom levels and image quality are inconsistent within and across the EcoUnits. However, one of the biggest challenges in the dataset is the occlusion caused by massive mutual overlaps of the plants, which is inherent to plant communities, irrespective of the number of co-occurring species. This makes a correct plant cover prediction very difficult.

While images from the EcoUnits have been collected daily, due to the necessity of annotating them, we can only utilize one image per week per camera for our investigations. This leads to only 682 images with annotations, which are vectors containing the estimated plant cover for each plant species of the dataset.

% Growth process in the images and reference to figure
% Compare contents with other paper

\textbf{Plant Cover Annotations.}
The plant cover is defined as the area of ground covered by each plant species. This value is typically estimated by a biologist directly in the field and can thus be highly subjective. Therefore, the annotations of the InsectArmageddon dataset, which were provided by Ulrich \etal \cite{ulrich2020invertebrate}, can be considered noisy. In addition to this, as an accurate percentage estimation is practically infeasible for humans, the estimated plant cover is usually done with a certain quantization, which in the case of this dataset is the so-called Schmidt-scale \cite{pfadenhauer1997vegetationsokologie}. The latter includes the percentages 0, 0.5, 1, 3, 5, 8, 10, 15, 20, 25, 30, 40, 50, 60, 70, 75, 80, 90 and 100 percent. The quantization process introduces additional label noise. It should be noted that, due to disregarding the occlusion during the estimation, the sum of the plant cover percentages can exceed 100\%.

\section{Method}
\label{sec:method}

We apply the same approach as in \cite{koerschens2020towards}: we view the plant cover prediction problem as a pixel-wise classification problem, where these classifications can then be aggregated into a final cover prediction. One additional advantage of this approach is that the pixel-wise classifications can be viewed as segmentation maps, serving as a way for the user to verify the correctness of the result.
Taking this approach requires us to identify the plant species at each location in the images before aggregating them into a single cover value for each species. \autoref{fig:examples_ia} shows that the plants are, in part, very similar to each other, and, due to the relatively large area captured in the single images, rather small compared to the whole image. In order to still be able to identify the plants correctly and be able to take into account their small and essential features, we need to process the images in a relatively high resolution. Often, when high-resolution images are used during CNN training, the images are divided into smaller parts, which are then used for training (e.g., \cite{he2017mask}). This approach, however, is not possible in our case, as the numerical plant cover annotations are only valid for the complete image. Therefore, we are required to train on high-resolution images directly.

While this was done already in our previous work \cite{koerschens2020towards} with a resolution of 672x336\pxns, here we test the effect of training on even higher resolutions. Hence, we evaluate our method on images with resolutions of 768x384 and 1536x768\pxns, respectively. To receive a better insight on the effect of different network architectures on the task of plant cover prediction, we also measure the performance of standard classification network architectures in conjunction with a Feature Pyramid Network (FPN) \cite{lin2017feature}. The networks we will compare are ResNet50 \cite{he2015resnet}, DenseNet121 \cite{iandola2014densenet}, and InceptionV3 \cite{szegedy2015inceptionv3}. ResNet50 was selected as it is used for comparison and as backbone in almost all possible settings and tasks, from simple classification \cite{he2015resnet} over object detection \cite{lin2017focal} to instance segmentation \cite{he2017mask}. InceptionV3, in turn, is commonly used in fine-grained classification tasks (e.g., \cite{cui2018large}), and DenseNet121 was chosen due to its particular network structure, which incorporates information from earlier layers directly in the decision process in later layers via long-range skip connections. As these networks only have output feature maps of small sizes, we also apply the FPN, which is usually utilized in tasks like object detection (e.g., \cite{ren2015faster,lin2017focal}) and semantic segmentation (e.g., \cite{he2017mask}), and could therefore be a sensible choice due to the similarity of the underlying task. The FPN increases the size of the output feature maps by incrementally fusing feature maps from the end of the network with higher resolution ones generated in earlier layers of the network. Hence, this structure also enables us to generate high-resolution output feature maps when joined with the abovementioned standard networks. This also allows us to evaluate whether generating higher resolution class outputs results in a better cover prediction, and to receive information on the usefulness of features from earlier layers.

As mentioned above, we also want to quantify the effect of in-domain and out-of-domain transfer learning. To this end, we use, for the in-domain pretraining, the previously introduced GBIF dataset, as due to the specifically selected plant species it is much more similar to our target task than, for example, the widely used ImageNet \cite{russakovsky2015imagenet} dataset containing merely general objects, which we use as out-of-domain pretraining for comparison.

%We found that the network proposed in \cite{koerschens2020towards} only performs poorly on such high resolution images, which is likely caused by the relatively small receptive field of the network.

% Motivation behind GBIF and ImageNet comparison
% Detection and segmentation references
% High resolution, comparison to other paper
% Also: Fine-grained problem with comparably small resolution
% Networks as solution for scaling up resolution for better identification results
% Reasoning FPN, selection of base networks in conjunction with pretraining

\subsection{Basic Network Structure}

\begin{figure}[t]
    \centering
    \includegraphics[width=0.8\textwidth]{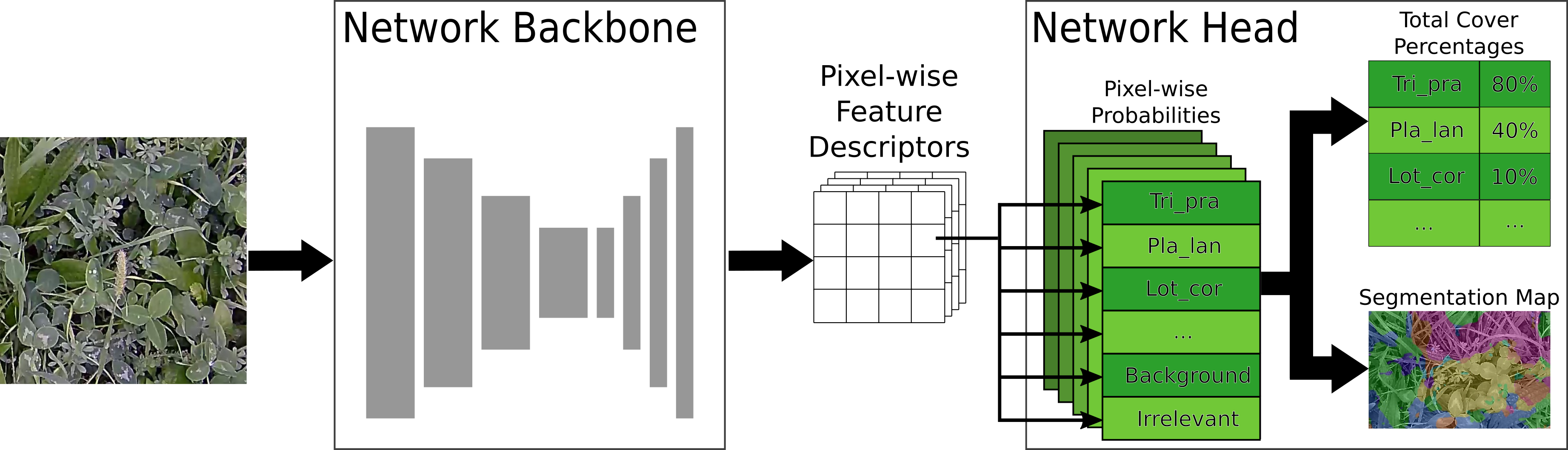}
    \caption{The basic structure of the network. The image is input in the network backbone, which extracts pixel-wise feature descriptors. These are then used by the network head to calculate the pixel-wise probabilities for each plant, which are afterwards aggregated into the total cover percentages. The generated pixel-wise probabilities can also be interpreted as a segmentation map, so that the predictions of the network can also be interpreted by the user.}
    \label{fig:network_structure}
\end{figure}

Our network consists of two main components: a feature extractor backbone and a network head. The backbone consists of one of the abovementioned standard classification networks potentially in conjunction with a Feature Pyramid Network to increase the output resolution. It hence generates a pixel-wise feature map, which is then used as input for the network head. The latter comprises a calculation model, explained in detail in \autoref{subsec:calculation_model}, which initially determines the pixel-wise probabilities of each plant species and afterward aggregates the predicted values into the total cover percentage of each species. As already mentioned, the generated class or probability map can also be used as a segmentation map for the user to confirm the correctness of the results. An overview of this network structure is shown in \autoref{fig:network_structure}.

% Feature extractor network + head
% Structure of feature extractor + FPN

\newcommand{\Aplant}{\ensuremath{A^{plant}}}
\newcommand{\Asoil}{\ensuremath{A^{uncovered}}}
\newcommand{\xy}{\ensuremath{_{x,y}}}
\newcommand{\p}{\ensuremath{^{p}}}
\newcommand{\Pplant}{\ensuremath{P\p\xy}}
\newcommand{\Abio}{\ensuremath{A^{bio}}}
\newcommand{\Abg}{\ensuremath{A^{bg}}}
\newcommand{\Airr}{\ensuremath{A^{irr}}}
\newcommand{\Atotal}{\ensuremath{A^{total}}}
\newcommand{\Pbio}{\ensuremath{P^{bio}\xy}}
\newcommand{\Pbg}{\ensuremath{P^{bg}\xy}}
\newcommand{\Pirr}{\ensuremath{P^{irr}\xy}}

\subsection{Calculation Model}
\label{subsec:calculation_model}

The calculation model represents the computations taking place in the network head to lastly generate the cover predictions. This calculation is separated into two parts: pixel-wise class predictions and aggregations.

\textbf{Class Prediction.}
As the pixel-wise predictions are largely dependent on the actual cover calculation, we will first introduce the intuitive formula to calculate the cover for a single plant species:

\begin{equation}
    cover^{plant} = \frac{\Aplant}{\Aplant + \Asoil}\:,
\end{equation}

where $\Aplant$ is the area covered by the single plant species and $\Asoil$ is the area not covered by it. Under this premise, there are several things to take into account for the pixel-wise class prediction. First, as occlusion is usually ignored for calculating the plant cover, it has to be possible to predict multiple different plant species in the same location. Second, the areas in which no plants exist also have to be taken into account in the calculation. Third, as seen in the example images in \autoref{fig:examples_ia}, there are areas in the image, which are irrelevant for cover calculation (e.g., the walls of the EcoUnits) and should thus be ignored.
While the existence of multiple plants in a single pixel does not have to be mutually exclusive, plants, soil and irrelevant pixels cannot be predicted in the same location and therefore have to be mutually exclusive. To be able to calculate the plant cover accurately, we also include these premises in our calculation model. As these factors are already included in the calculation model from our previous work \cite{koerschens2020towards}, we also utilize this model here with slight modifications.

We define $\Pplant$ as the probability of a plant $p$ existing at location $x,y$. We also define $\Pbio$, $\Pbg$, and $\Pirr$ as the probability of any plant, background (uncovered soil), and irrelevant data being at location $x,y$, respectively. Note that, as mentioned above, the single plant probabilities $\Pplant$ do not necessarily add up to one, as they are not mutually exclusive, while the three probabilities $\Pbio$, $\Pbg$ and $\Pirr$ do.

As $\Pbio$ is naturally dependent on the probabilities $\Pplant$, we can set them in relation, as in \cite{koerschens2020towards}:

\begin{equation}
    \Pbio = 1 - (\Pbg + \Pirr) = 1 - \frac{\kappa}{\kappa + \sum_p \Pplant} = \frac{\sum_p \Pplant}{\kappa + \sum_p \Pplant}\:,
\end{equation}

where $\kappa$ is a threshold value, which determines if a pixel is considered a plant pixel or not. While $\kappa$ is a tunable hyperparameter in \cite{koerschens2020towards}, we use it as a learnable parameter in our experiments. 

\textbf{Aggregation.} Now, to aggregate the predicted probabilities into the plant-wise cover percentages, we can calculate the areas 

\begin{align}
    (\Abg, \Airr, \Abio) &= (\sum\xy \Pbg, \sum\xy \Pirr, \sum\xy \Pbio)\:,\\
    \Atotal &= \Abio + \Abg + \Airr\:.
\end{align}
%\begin{align}
%    \Abg &= \sum\xy \Pbg\:,\\
%    \Airr &= \sum\xy \Pirr\:,\\
%    \Abio &= \sum\xy \Pbio\:,\\
%    \Atotal &= \Abio + \Abg + \Airr\:.
%\end{align}
Thus, we can finally calculate the cover values for each plant $p$:

\begin{equation}
    cover\p = \frac{\sum\xy \Pplant}{\Atotal - \Airr} = \frac{\sum\xy \Pplant}{\Abio + \Abg}
\end{equation}

While in \cite{koerschens2020towards} we also applied a vegetation percentage, i.e., the percentage of ground covered by plants in general, as an auxiliary value, we found that this is not necessary when using pretrained networks, which in turn removes the need for additional annotations.

% Explanation of calculation model and modifications compared to other paper

\section{Experiments}
\label{sec:experiments}

In our experiments, we compare the performance of the previously introduced network architectures with each other using two different resolutions. This comparison is done first using in-domain pretraining for each of the networks and then repeated using ImageNet pretraining. At the end of this section, we provide an error analysis based on the experiment duration of the InsectArmageddon dataset.
The metrics we will use in our experiments for performance evaluation are the mean absolute error (MAE) and the mean scaled absolute error (MSAE) as defined in \cite{koerschens2020towards}. As the MAE is heavily influenced by the average cover percentages of each plant species, we introduced the MSAE, which is the mean absolute error divided by the species-wise plant cover mean, to allow for a fairer comparison between the different species. For the exact values used to scale the errors, we would like to refer the reader to \cite{koerschens2020towards}.

\subsection{Comparison of Different Networks}
\label{subsec:exp1}

In this experiment different network architectures were utilized as backbone and their performance is compared. Three different architectures were used: ResNet50 \cite{he2015resnet}, InceptionV3 \cite{szegedy2015inceptionv3} and DenseNet121 \cite{iandola2014densenet}.
We applied these networks in two different versions: with and without an FPN. In the latter case, we utilized an FPN \cite{lin2017feature} with depth 512 to increase spatial resolution and hence to be able to better predict the classes pixel-wise. We used the P2 layer of the FPN, i.e., the second-to-last layer of the FPN, resulting in a final resolution of $\frac{1}{4}$th of the original input image resolution.

\paragraph{Experimental Setup.}

\let\subsectionautorefname\sectionautorefname
\let\subsubsectionautorefname\sectionautorefname

In the following experiments, pretraining on the GBIF dataset (see \autoref{subsec:gbif}) was utilized, whereafter the main training on the InsectArmageddon (\autoref{subsec:ia_ds}) dataset took place.

The pretraining on GBIF was done on the base network and the FPN, if used. All networks were initialized with off-the-shelf ImageNet weights provided by the Keras framework \cite{chollet2015keras}. During the pretraining we used an image resolution of $448\times448$\px for ResNet50 and DenseNet121, and $427\times427$\px for InceptionV3 due to its different network structure. Each network was trained using a batch size of 12 for 20 epochs with an initial learning rate of 0.01, decaying by a factor of 0.1 at epoch 10 and 15. To improve the adaptability of the learned weights on the InsectArmageddon dataset, we also utilized several data augmentations to improve generalization and adaptation to the target dataset: random rotations, random crops, and random horizontal flipping. The optimizer during training is SGD with a momentum of 0.9, and the loss is the standard categorical cross-entropy.

During training on the InsectArmageddon dataset, we exchange the classification head of the network from the pretraining step with the plant cover prediction head explained in \autoref{subsec:calculation_model}, which we then train in conjunction with the FPN, if used. The rest of the network is frozen during training, as in previous experiments not shown here we found that fully fine-tuning the network does either not change the results or is even detrimental to them.
We also found that for this training, a batch size of 1 is sufficient, and larger batch sizes also do not change the results. We assume this is the case due to the calculation performed in the network head. The latter can be interpreted as performing many single classifications with strongly changing and weakly defined target values. This, in turn, massively regularizes the network to a point at which bigger batches and additional regularization do not improve the results in any way.
The networks are trained using the Adam \cite{Kingma2015AdamAM} optimizer for faster convergence for 40 epochs starting with a learning rate of 0.001 (0.00001 in case of the network containing an FPN), decaying with a factor of 0.1 at epoch 20 and 30. We apply the mean absolute error as loss during the training process. Only horizontal flips are applied as data augmentation, as we found that the effect of most augmentations was either negligible or detrimental to the results.
All experiments are performed in a special 12-fold cross-validation, i.e., during each fold, we select two EcoUnits, whose images are used for testing, while the images of the remaining units are used for training. For generating the experimental results below, the results of these folds are averaged.

\paragraph{Results.}

% \begin{table}[t]
%     % Experiments without occlusion augmentations, but with rotation
%     \centering
%     \caption{The numerical results of the experiment using several standard networks with pretraining on the GBIF dataset. Top results are marked in bold font. MAE and MSAE represent the \textit{mean absolute error} and the \textit{mean scaled absolute error}, respectively.}
%     \begin{tabular}{l|c|c|c|c}
%         \textbf{Image Resolution} & \multicolumn{2}{c|}{\textbf{$768\times384$}} & \multicolumn{2}{c}{\textbf{$1536\times768$}} \\\hline
%         \textbf{Metric} & \textbf{MAE} & \textbf{MSAE} & \textbf{MAE} & \textbf{MSAE} \\\hline
%     ResNet50 \cite{he2015resnet} & 5.32\% & 0.515 & 5.18\% & 0.502 \\
%     ResNet50 \cite{he2015resnet} + FPN \cite{lin2017feature} & 5.53\% & 0.529 & 5.26\% & 0.501 \\\hline
%     InceptionV3 \cite{szegedy2015inceptionv3} & 5.67\% & 0.541 & 5.23\% & 0.505 \\
%     InceptionV3 \cite{szegedy2015inceptionv3} + FPN \cite{lin2017feature} & 5.53\% & 0.527 & 5.40\% & 0.520 \\\hline
%     DenseNet121 \cite{iandola2014densenet} & 5.26\% & 0.507 & \textbf{5.14}\% & \textbf{0.492} \\
%     DenseNet121 \cite{iandola2014densenet} + FPN \cite{lin2017feature} & 5.41\% & 0.520 & 5.30\% & 0.500 \\      
%     \end{tabular}
%     \label{tbl:results_gbif}
% \end{table}

\begin{table}[t]
    % Experiments without occlusion augmentations, but with rotation, always trained FPN
    \centering
    \caption{The numerical results of the experiment using several standard networks with pretraining on the GBIF dataset. Top results are marked in bold font. MAE and MSAE represent the \textit{mean absolute error} and the \textit{mean scaled absolute error}, respectively.}
    \begin{tabular}{l|c|c|c|c}
        \textbf{Image Resolution} & \multicolumn{2}{c|}{\textbf{$768\times384$}} & \multicolumn{2}{c}{\textbf{$1536\times768$}} \\\hline
        \textbf{Metric} & \textbf{MAE} & \textbf{MSAE} & \textbf{MAE} & \textbf{MSAE} \\\hline
    ResNet50 \cite{he2015resnet} & 5.49\% & 0.528 & 5.22\% & 0.503 \\
    ResNet50 \cite{he2015resnet} + FPN \cite{lin2017feature} & 5.35\% & 0.512 & 5.18\% & 0.496 \\\hline
    InceptionV3 \cite{szegedy2015inceptionv3} & 5.51\% & 0.531 & 5.30\% & 0.506 \\
    InceptionV3 \cite{szegedy2015inceptionv3} + FPN \cite{lin2017feature} & 5.51\% & 0.528 & 5.27\% & 0.503 \\\hline
    DenseNet121 \cite{iandola2014densenet} & 5.37\% & 0.522 & 5.18\% & 0.495 \\
    DenseNet121 \cite{iandola2014densenet} + FPN \cite{lin2017feature} & 5.30\% & 0.515 & \textbf{5.16}\% & \textbf{0.494} \\      
    \end{tabular}
    \label{tbl:results_gbif}
\end{table}

The results of the GBIF-pretraining experiments can be seen in \autoref{tbl:results_gbif}. It is visible that the best results concerning the mean absolute error (MAE) and mean scaled absolute error (MSAE) were achieved with a DenseNet121 with FPN. We can also note that the DenseNet outperforms the two other networks in almost every single setting, which can likely be attributed to its structure, that directly incorporates information from earlier layers in the network for the final prediction. Similarly, the ResNet50 outperforms InceptionV3 in all settings, which leads to the assumption that the structure of the Inception modules might be detrimental in this task.
From these results, we can also observe that using a higher resolution is strictly better than using the lower one, which means that, even though the $768\times384$\px resolution is already quite high, there is still more relevant information contained in the images, which can only be utilized when using an even higher image resolution.
In the experimental data, we can also see that the results improve upon using the FPN for some networks more than for others. The network profiting most from the FPN is the ResNet50, which seems to produce useful information for identification at multiple levels of the FPN and therefore receives a substantial performance boost. The DenseNet, as mentioned before, already includes low-level features by default in its network structure, which is why the FPN does not add much novel information and therefore only improves results slightly. While the InceptionV3 does not have long-range skip connections, the improvements by addition of the FPN are also only minor. While the reason for this is not entirely apparent, this could be caused by the differing network structure as compared to the ResNet50 and DenseNet121, which might not benefit as much from the FPN.
The top result of the DenseNet121 outperforms the network proposed in \cite{koerschens2020towards} in terms of mean absolute error (5.16\% vs. 5.30\%) and mean scaled absolute error (0.494 vs. 0.500). However, it should also be noted that in this work, we completely abstain from using the auxiliary vegetation-percent annotations, which were used in \cite{koerschens2020towards} in addition to the plant cover annotations, and the amount of training necessary here is also much less, while still outperforming the network from \cite{koerschens2020towards}. Additionally, it should be noted that, while the differences in errors are comparably small, visual inspection suggests that this is likely primarily due to improvements in the detection of less abundant plants, which have a rather small influence on the total error value.

\subsection{Comparison with ImageNet Pretraining}

\let\subsectionautorefname\sectionautorefname
\let\subsubsectionautorefname\sectionautorefname

 To evaluate the effect of the additional in-domain pretraining using the GBIF dataset in \autoref{subsec:exp1} we conduct an ablation study and compare it with a standard ImageNet pretraining. It should be noted that pretrained ImageNet weights were only available for the base classification networks without FPN, which is then randomly initialized. The pretraining on the GBIF dataset in \autoref{subsec:exp1} has been performed on the base network including the FPN.

\paragraph{Experimental Setup.}

In these experiments, the setup is very similar to the one in the experiments before. However, instead of pretraining the network on GBIF, we merely load off-the-shelf ImageNet weights for the base networks, which are provided by the Keras framework \cite{chollet2015keras}. During training, in case of not using an FPN, we again only train the last layer using the same scheme as described above with an initial learning rate of 0.001. In the cases where an FPN is used, as its weights are not initialized, we train the FPN and the cover prediction head jointly with a learning rate of 0.00001.

\paragraph{Results.}

\begin{table}[t]
    \centering
    \caption{The numerical results of the experiment using several standard networks with only ImageNet pretraining. Top results are marked in bold font. MAE and MSAE represent the \textit{mean absolute error} and the \textit{mean scaled absolute error}, respectively.}
    \begin{tabular}{l|c|c|c|c}
        \textbf{Image Resolution} & \multicolumn{2}{c|}{\textbf{$768\times384$}} & \multicolumn{2}{c}{\textbf{$1536\times768$}} \\\hline
        \textbf{Metric} & \textbf{MAE} & \textbf{MSAE} & \textbf{MAE} & \textbf{MSAE} \\\hline
    ResNet50 \cite{he2015resnet} & 5.55\% & 0.537 & 5.32\% & 0.510 \\
    ResNet50 \cite{he2015resnet} + FPN \cite{lin2017feature} & 5.37\% & 0.521 & 5.30\% & 0.510 \\\hline
    InceptionV3 \cite{szegedy2015inceptionv3} & 5.74\% & 0.546 & 5.34\% & 0.505 \\
    InceptionV3 \cite{szegedy2015inceptionv3} + FPN \cite{lin2017feature} & 5.62\% & 0.535 & 5.33\% & 0.506\\\hline
    DenseNet121 \cite{iandola2014densenet} & 5.39\% & 0.515 & \textbf{5.17}\% & \textbf{0.491} \\
    DenseNet121 \cite{iandola2014densenet} + FPN \cite{lin2017feature} & 5.31\% & 0.510 & 5.20\% & 0.494 \\      
    \end{tabular}
    \label{tbl:results_imagenet}
\end{table}

The results of our experiments with off-the-shelf ImageNet weights can be seen in \autoref{tbl:results_imagenet}. We can see that the top results are generated again by the DenseNet121, with an MAE of 5.17\% and an MSAE of 0.491. These results differ only slightly from the previous experiments. This difference likely results from the structure of the DenseNet, which directly incorporates features from earlier layers into top-level predictions, which are beneficial for high-resolution plant predictions. Here, we can also see that the FPN appears to be beneficial in all cases for the lower image resolution. For the higher resolutions, we notice that the performances of the networks with FPN are very similar to the ones of the standalone base networks in all cases. This leads to the assumption that the learned lower-level features incorporated by the FPN are not very relevant on this resolution anymore, and higher-resolution ones become more crucial. Compared to the GBIF pretraining experiments, we can see that if we only consider the base networks without FPN, the GBIF pretraining is beneficial in almost all cases, with only a slight deviation from this pattern for DenseNet121.

\subsection{Error Analysis}

\begin{figure}[t]
    \centering
    \begin{subfigure}{0.47\linewidth}
        \centering
        \includegraphics[width=\linewidth]{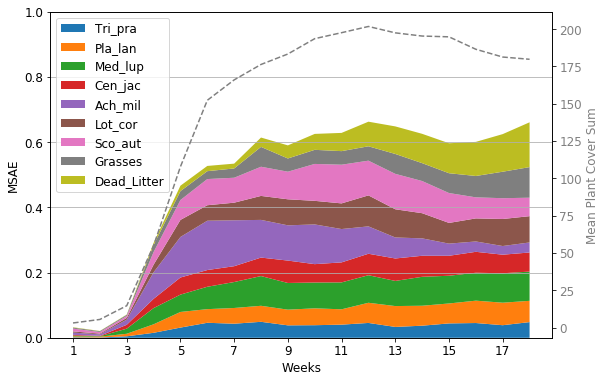}        
    \end{subfigure}
    \begin{subfigure}{0.43\linewidth}  
        \centering      
        \includegraphics[width=\linewidth]{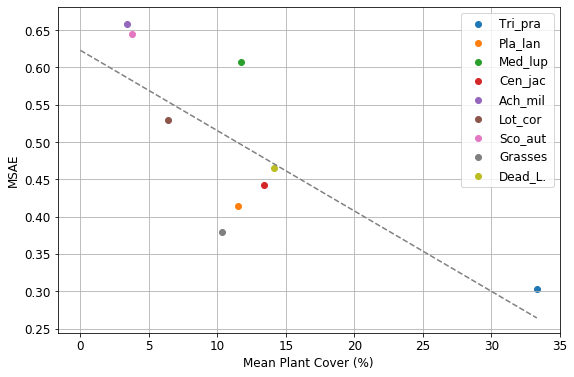}  
    \end{subfigure}
    \caption{Left: The mean scaled absolute errors (MSAE) averaged over all plant species in the dataset in dependence on the week of the image recording. For reference, the plant cover sum averaged over all weeks is shown as dashed line. Right: The relationship between the species-wise mean plant cover and the mean scaled absolute error (MSAE). We can see that the values are not or even negatively correlated, i.e., as opposed to the MAE the MSAE does not depend strongly on the plant abundances (correlation of $p=-0.774$; $R^2=0.600$).
    }
    \label{fig:msae_weekbased_full}
\end{figure}
\begin{figure}[t]
    \centering
    \includegraphics[width=0.67\textwidth]{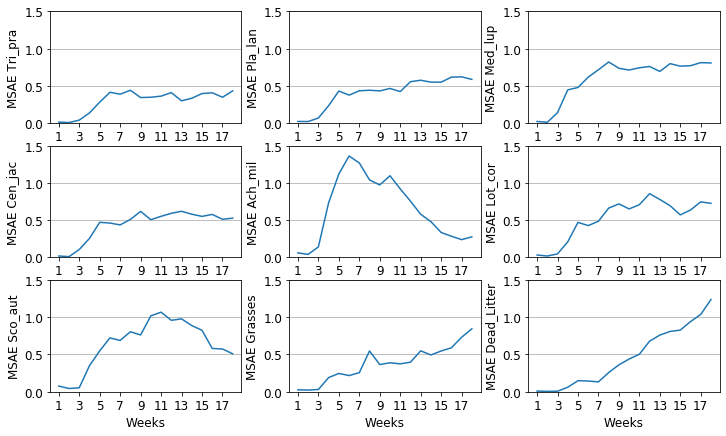}
    \caption{The mean scaled absolute error in dependence on the week of the image recording.}
    \label{fig:msae_weekbased_plants}
\end{figure}
%\begin{figure}[t]
%    \centering
%    \includegraphics[width=0.45\textwidth]{images/diagrams/resnet_gbif_1536_mae_plantwise.png}
%    \includegraphics[width=0.45\textwidth]{images/diagrams/resnet_gbif_1536_msae_plantwise.png}
%    \caption{The distribution of absolute errors (left) and of scaled absolute errors (right) of each plant species in the dataset.}
%    \label{fig:mae_msae_plantwise}
%\end{figure}

To finally also analyze the output and errors of our method, we take a look at the error of the DenseNet121 pretrained on GBIF with FPN for an input image resolution of $1536\times784$\px from \autoref{subsec:exp1} plotted over the weeks recorded in the dataset. As mentioned above, the images contain time series of images collected weekly over a range of 18 weeks, with the plants growing progressively over time while dying towards the end of the experiment. As these changing conditions can hugely affect the outcome of the automatic analysis, we observe the error of our method over the single weeks. Here, we will only focus on the MSAE.
An overview of the MSAE changing over time can be seen in \autoref{fig:msae_weekbased_full}, in which the change of the average plant cover percentage summed over all plants is also shown for comparison. We see that in the first few weeks, where not many plants have grown, the error is small but starts rising strongly around week 4, which coincides with the plant growth. Following this sudden rise, the error stays mostly consistent, also coinciding with the cover percentages. This high error can probably be attributed to the large amount of occlusion taking place in the images. It is visible in the graph that the latter peaks at 200\%, meaning about half of the plant area used for estimating the ground truth values is occluded, making a correct estimation for the network difficult. Towards the end of the experiment, we can see that the error does not coincide with the changes in cover percentages anymore, as the former is rising while the latter are dropping. This can be explained when taking a look at the plant-wise errors in \autoref{fig:msae_weekbased_plants}. There we can see that the error for most plants is decreasing, as they are dying towards the end of the experiment. This process, i.e., the death of plants by either age (senescence) or due to the experimental setup, leads to plants being counted not as their original class anymore, but as the \textit{Dead litter} class, where we see a rise in the error. This creates the assumption that the network cannot really identify the \textit{Dead litter} correctly and possibly still assigns the respective pixels to its original class. From this, we can assume that the network could not work out a suitable discrimination between \textit{Dead litter} and the other classes. This is understandable, as the point in time at which a plant starts to be considered \textit{Dead litter} can be quite arbitrary and subjective. While its label changes over the duration of one week, the plants might still bear significant resemblances to their previous label.
In summary, we can therefore say that most of the errors are likely caused by the amount of occlusion in the images and the problem of discriminating between the regular plant classes and \textit{Dead litter}. This, however, could also be caused by the non-existence of the \textit{Dead litter} class in the pretraining dataset.

% Maybe confirm last sentence on basis of ImageNet pretraining?

%As stated above, our method, like the one from \cite{koerschens2020towards}, also generates an implicitly learned segmentation map. In this section we want to present a selection of example maps generated during the prediction process.

\section{Conclusions and Future Work}
\label{sec:conclusion}

In this work, we found that utilizing standard CNNs can yield similar or slightly better results than using, for example, our previously proposed network from \cite{koerschens2020towards}. We also found that the usage of FPNs is in general beneficial, while the ResNet50 benefits most from them. It should also be noted that FPNs increase the output resolution of the resulting segmentation map, which, in turn, leads to better interpretability of the results compared to their non-FPN counterparts. We also found that using higher image resolutions is always beneficial in our setting. Therefore, collecting images with high resolutions should be prioritized to improve prediction accuracy. Pretraining the network on a more related dataset rather than using off-the-shelf ImageNet weights proved beneficial for all the networks with only few exceptions when using DenseNet121. Hence an in-domain pretraining should, in most cases, be preferred to an out-of-domain one, if available. Also, with pretraining, we can achieve similar or better results to \cite{koerschens2020towards} with only training a part of the network on the final task, saving time and hardware capacities for training on the plant cover prediction task.
From our week-based error analysis, we were able to conclude that some parts of the errors are most likely caused by the massive occlusions in the images, as well as the network's inability to differentiate between \textit{Dead litter} pixels and its preceding plant species before its death.

In future work, we aim to tackle the abovementioned problems. For occlusion, some approaches in the area of amodal segmentation \cite{li2016amodal} might be suitable. For a better differentiation between plants, other pretraining datasets like iNaturalist \cite{vanhorn2017inaturalist} or even pretraining the network on segmentation datasets like MSCOCO \cite{lin2014microsoft} to prime it for pixel-wise classification might be a possibility.

\section*{Acknowledgement}
Matthias Körschens thanks the Carl Zeiss Foundation for the financial support. We would also like to thank Alban Gebler and the iDiv for providing the data for our investigations.

% FPN increases resolution, rarely improves results
% ResNet50 with higher resolution performs best
% Higher resolution always beneficial
% In total: results rather similar to previous approach
% Less training time + no vegetation annotations needed

% Pretraining on similar dataset somewhat effective, ImageNet pretraining worse than previous approach

% Errors get higher towards later weeks, mostly caused by occlusion and dead litter (cannot be differentiated from other plants)

% Future work: Approaches to tackle occlusion, alternative pretraining approaches for improved segmentation

%%% Angabe der .bib-Datei (ohne Endung) / State .bib file (for BibTeX usage)
\bibliography{references} %\printbibliography if you use biblatex/Biber
\end{document}